\def\editmode{0}
\def\reportmode{0}

\def\bibfilenames{WISENET,Hyperparams}

\if\reportmode1
\documentclass[10pt,final,onecolumn]{IEEEtran}
\else
\documentclass[conference]{IEEEtran}
\IEEEoverridecommandlockouts
\fi

\if\editmode1  
\usepackage[backend=bibtex,style=alphabetic,sorting=debug]{biblatex}
\DeclareFieldFormat{labelalpha}{\thefield{entrykey}}
\DeclareFieldFormat{extraalpha}{}
\bibliography{\bibfilenames}

\newcommand{\cmt}[1]{\noindent\textcolor{lightgreen}{\underline{[#1]}}} 
\newenvironment{myitemize}{\begin{itemize}}{\end{itemize}}
\newcommand{\myitem}{\item}

\newcommand{\acom}[1]{\textcolor{red}{[#1]}}

\else
\usepackage{cite}
\newcommand{\cmt}[1]{} 
\newenvironment{myitemize}{}{}
\newcommand{\myitem}{}

\newcommand{\acom}[1]{}

\fi 

\usepackage{fixltx2e}

\usepackage{graphicx}

\usepackage{subcaption}              

\usepackage[utf8]{inputenc}

\usepackage{amsfonts}
\usepackage{amsmath}
\usepackage{mathtools}

\usepackage{amssymb}

\usepackage{bm}

\usepackage{color,verbatim}
\usepackage{multirow}
\usepackage{accents}

\usepackage{theoremref}

\usepackage{url}

\newcounter{rulecounter}
\newcommand{\resetrule}{ \setcounter{rulecounter}{0}}
\resetrule

\newsavebox{\selvestebox}
\newenvironment{colbox}[1]
  {\newcommand\colboxcolor{#1}%
   \begin{lrbox}{\selvestebox}%
   \begin{minipage}{\dimexpr\columnwidth-2\fboxsep\relax}}
  {\end{minipage}\end{lrbox}%
   \begin{center}
   \colorbox{\colboxcolor}{\usebox{\selvestebox}}
   \end{center}}

\definecolor{orange}{rgb}{1,0.8,0}
\definecolor{gray}{rgb}{.9,0.9,0.9}
\definecolor{darkgray}{rgb}{.3,0.3,0.3}
\definecolor{darkblue}{rgb}{.1,0.0,0.3}
\definecolor{lightblue}{rgb}{0.7,0.7,1}
\definecolor{lightred}{rgb}{1,0.7,.7}
\definecolor{purple}{RGB}{204,153,255}
\definecolor{lightgray}{rgb}{.95,0.95,0.95}
\definecolor{lightgreen}{rgb}{0.3,0.5,0.3}
\definecolor{darkgreen}{rgb}{0.05,0.3,0.05}

\newtheorem{myproposition}{Proposition}
\newtheorem{myremark}{Remark}
\newtheorem{myproblemstatement}{Problem Statement}
\newtheorem{mylemma}{Lemma}
\newtheorem{mytheorem}{Theorem}
\newtheorem{mydefinition}{Definition}
\newtheorem{mycorollary}{Corollary}

\usepackage{mathtools}
\usepackage{bbold}
\usepackage{algorithm}
\usepackage[noend]{algpseudocode}
\newcommand{\remove}[1]{}
\usepackage{epstopdf}

\begin{document}

\title{Online Hyperparameter Search Interleaved with Proximal Parameter Updates}

\if\reportmode1
  \author{Luis Miguel Lopez-Ramos and Baltasar Beferull-Lozano\\[.5cm]\today}
\else
\author{Luis M. Lopez-Ramos,~\IEEEmembership{Member,~IEEE,} and Baltasar Beferull-Lozano,~\IEEEmembership{Senior Member,~IEEE}
\thanks{This work was supported by grants SFI Offshore Mechatronics 237896/E30, PETROMAKS Smart-Rig 244205/E30, IKTPLUSS INDURB 270730/O70}
\thanks{The  authors  are  with  the  WISENET  Center,  Dept.  of  ICT,  University  of Agder,  Jon  Lilletunsvei  3,  Grimstad,  4879  Norway.  E-mails:\{luismiguel.lopez, baltasar.beferull\}@uia.no.}
}
\fi

\maketitle
\begin{abstract}
There is a clear need for efficient algorithms to tune hyperparameters for statistical learning schemes, since the commonly applied search methods (such as grid search with N-fold cross-validation) are inefficient and/or approximate. Previously existing algorithms that efficiently search for hyperparameters relying on the smoothness of the cost function cannot be applied in problems such as Lasso regression. 
In this contribution, we develop a hyperparameter optimization method that relies on the structure of proximal gradient methods and does not require a smooth cost function. Such a method is applied to Leave-one-out (LOO)-validated Lasso and Group Lasso to yield efficient, data-driven, hyperparameter optimization algorithms.
Numerical experiments corroborate the convergence of the proposed method to a local optimum of the LOO validation error curve, and the efficiency of its approximations.

\end{abstract}

\if\reportmode0
\fi

\section{Introduction}

\cmt{overview} Given their proven utility to control the model complexity, hyperparameters are crucial for a successful application of many statistical learning schemes in real-world engineering problems. The generalization capability and performance of such schemes on unknown instances can be improved with a careful hyperparameter selection. 
\cmt{traditionally} Regularized models control the trade-off between a data fidelity term and a complexity term known as regularizer by means of one or several hyperparameters. Ridge regression, Lasso, Group Lasso, and Elastic net are instances of regularized models. While the regression weights can be optimized efficiently via the  proximal gradient descent (PGD) method and its variants, the associated hyperparameter optimization (HO) is a non-convex, challenging problem \cite{bergstra2011hyper}.  
\cmt{motivation-causality} One main motivation to develop HO schemes for PGD-based learning algorithms is the interest in solving for models with sparsity, which can enhance their interpretability.

\cmt{LOO, CV, grid and random search} Given a dataset in batch form, a commonly applied criterion for hyperparameter optimization is the leave-one-out (LOO) validation error, because it reflects the ability of an estimator to predict outputs for unobserved patterns \cite{homrighausen2014leave}. 
The computational cost of evaluating the LOO validation error grows superlinearly with the number of data points, so that it is often approximated by N-fold cross validation (CV) with a small N (e.g., 10). Common practice to search for (sub)optimal hyperparameters is to use grid search or random search \cite{bergstra2011hyper,bergstra2012random} because of their simplicity.

\cmt{config evaluation} An improved form of random search are \emph{configuration-evaluation} methods, which focus the computation resources in promising hyperparameter configurations by quickly eliminating poor ones, important examples of it being the Hyperband \cite{li2018hyperband} and Bayesian optimization-based approaches in \cite{klein2017fast}.

\cmt{Prior work}
\begin{myitemize}
\myitem \cmt{smooth functions} Gradient-based (exact and approximate) HO methods have been proposed recently for problems where the cost function is smooth. Several recent approaches formulate a bi-level program where an inner program is the optimization of the model parameters (model weights in the case of regression) and the outer program is the minimization of a surrogate of the generalization capability (e.g. validation MSE). In particular, \cite{pedregosa2016approximate} applies the implicit function Theorem to a stationarity condition to obtain the hypergradient (gradient of the outer cost function w.r.t. the hyperparameters); however, this approach requires calculating the Hessian w.r.t. the model parameters and, consequently, it cannot be applied to widely used non-smooth regularizers (such as Lasso/group Lasso).

The approaches in \cite{monti2018adaptive,franceschi2017forward,franceschi2018bilevel} obtain a hypergradient by modeling the optimization of the regression weights as a dynamical system, where the state space is the parameter space and each iteration corresponds to a mapping from/to the same space. While \cite{monti2018adaptive} requires the aforementioned mapping to be invertible, \cite{franceschi2017forward,franceschi2018bilevel} avoid such a requirement by resorting to an approximation. This work combines ideas from \cite{pedregosa2016approximate,franceschi2017forward} to formulate a different implicit equation, derive the exact hypergradient, and develop a method that can work with non-smooth regularizers and, additionally, admits an online variant.

\myitem \cmt{sequential data} If the data is received in a streaming fashion, and the data distribution is time-varying, one may be interested in algorithms that find the right regularization parameter in different time segments or data windows, such as the adaptive approach in \cite{monti2018adaptive}, which is specific for Lasso estimators. On the contrary, our approach is general enough to be applied to several generalizations of Lasso, such as Group Lasso.

\myitem \cmt{hypernetwork} On the other hand, methods that train a neural network to predict optimal regression weights given a hyperparameter vector \cite{lorraine2018stochastic} have been proposed, together with approximations that alternate between updating the neural network weights and the hyerparameters. However, these methods incur in heavy over-parameterization, to the point of requiring more neural network parameters than the dimensionality of the regression weights and the hyperparameters together. 

\myitem \cmt{approximate Leave-one-out} Another approximation alleviating computation in hyperparameter search is porposed in \cite{wang2018approximateloo}, where the structure of specific estimators such as Lasso is exploited to approximately compute the LOO error metric at a very low cost. Note the difference with \cite{lorraine2018stochastic} because here it is only the error metric what is approximated, instead of the parameter vector. Despite the reduced computation, using this approximation for HO still requires a grid/random search scheme, which does not scale well with the dimensionality.

\end{myitemize}

\cmt{aim of this paper} In this paper, we propose and evaluate a method that jointly optimizes the regression weights and hyperparameter of a (Group-) Lasso regression model and converges to a stationary point of the LOO error curve. Our method can be extended to other estimators with proximable, non-smooth regularizers. The formulation is inspired by the forward-mode gradient computation in \cite{franceschi2017forward}, but where we use efficient approximations based on online (stochastic) gradient descent.

\cmt{contributions} 
The contributions and structure of the present paper are listed in the following:
\begin{myitemize}
\myitem Sec. \ref{sec:formulation} provides the general formulation for the HO in supervised learning and presents the use of PGD for our problem. 
\myitem In Sec. \ref{sec:hypergradient}, we present the derivation of the hypergradient (gradient w.r.t the hyperparameters). 
\myitem In Sec. \ref{sec:nonsmooth}, we discuss how to design our method for non-smooth cost functions in problems such as Lasso and Group Lasso. 
\myitem The main contribution is presented in Sec. \ref{sec:approximate}, consisting in the derivation of an online algorithm and an approximate scheme, both aimed at saving computation. 
\myitem Sec. \ref{sec:numeric} contains numerical tests with synthetic data, and concludes the paper.
\end{myitemize}

\section{Problem formulation}\label{sec:formulation}

Given a set of training input/label pairs $\{x_i, y_i\}_{i=1}^N$, with $x_i \in \mathbb{R}^P$ and $y_i \in\mathbb{R}$, consider the supervised learning problem of minimizing a linear combination of empirical risk (data fit) and structural risk (regularization term):
\begin{equation}\label{eq:estimation}
    {w}^\ast( \lambda, \mathcal{B}) := \arg \min_w \frac{1}{|\mathcal{B}|}\sum_{i \in \mathcal{B}} \ell_i(w) + \lambda^\top \Omega(w),
\end{equation}
for $\lambda \in \mathbb{R}_+^D$. This can be for instance particularized to the Lasso regression problem with $w \in \mathbb{R}^P$, $\ell_i(w) = (y_i - x_i^\top w)^2$, and $\Omega(w) = \|w \|_1$; section \ref{sec:nonsmooth} discusses other estimators.

It is well known that minimizing the empirical risk (in-sample error) does not guarantee that the estimated model will predict labels of unobserved inputs with low error. The role of regularization is to select the right model complexity, and the right choice of the hyperparameter $\lambda$ is crucial. To this end, any estimator in the form \eqref{eq:estimation} can be embedded in the bi-level optimization problem (minimization of the validation error):
\begin{equation} \label{eq:bilevel}
    \breve{\lambda}^\ast := \arg \min_\lambda \frac{1}{|\mathcal{V}|} \sum_{j \in \mathcal{V}} \ell^{\mathrm{VAL}}_j(w^\ast(\lambda, \mathcal{B}_j)).
\end{equation}
where $\mathcal{V}$ denotes the set of validation samples, and $\mathcal{B}_j$ denotes the training batch associated with the $j$-th validation sample. A typical choice in supervised learning is $\ell^{\mathrm{VAL}}_j(w) = (y_j - x_j^\top w)^2$. Since \eqref{eq:bilevel} may have several local minima, the notation $\breve{\lambda}^\ast$ is reserved for a global minimizer, whereas ${\lambda}^\ast$ will be used throughout the text to denote a stationary point.

Regarding the collection of training batches and the validation samples:
\cmt{examples}
\begin{myitemize}
\myitem In a held-out validation scheme, $\mathcal{B}_j = \mathcal{B}\; \forall j$, and 
$\mathcal{V} \cap \mathcal{B} = \emptyset$.
\myitem In $N$-fold cross-validation (CV), $\mathcal{V}$ is the train-and-validate dataset; the {\em folds} $\{ \mathcal{F}_1, \dots, \mathcal{F}_N \}$ are a partition of $\mathcal{V}$; and $\mathcal{B}_j = \bigcup_{j \notin \mathcal{F}_n} \mathcal{F}_n$.
\myitem {\em Leave-one-out} (LOO) validation is a special case of CV where $N= |\mathcal{V}|$, and $ \mathcal{F}_i = \{i \} \; \forall \; i$; and therefore, $\mathcal{B}_j = \mathcal{V} \setminus \{ j \}$.
\end{myitemize}


The rest of this section reviews how $w^\ast(\lambda, \mathcal{B}_j)$ is obtained. The next section will discuss the minimization of~\eqref{eq:bilevel} via the computation of the gradient w.r.t. the hyperparameter $\lambda$, also referred to as \emph{hyper-gradient}~\cite{maclaurin2015gradient,franceschi2017forward}.

\subsection{Proximal Gradient Descent}

The proximal gradient descent (PGD) algorithm allows to iteratively compute $w^\ast(\lambda, \mathcal{B}_j)$ given the training batch $\mathcal{B}_j$ and the hyperparameter $\lambda$, and it is advocated here for its simplicity. Extending our formulation to accommodate algorithms such as the accelerated PGD (which gives rise to FISTA when applied to $\ell_1$-regularized problems) is out of the scope of the present paper and left as future work. 

Given a function $\Psi$, the proximity (prox) operator is defined as \cite{parikh2014proximal}	
\begin{equation} \label {eq:defproximaloperator}
	\mathrm{prox}_{ \Psi}^{\eta}(\bm v)
	\triangleq 
	\underset{\bm x \in \text{dom }\Psi}{\arg\min}\left [\Psi(\bm x)+\frac{1}{2\eta}\left \lVert \bm x-\bm v\right \rVert_2^2\right].
\end{equation}

If $\Omega$ is such that the prox operator can be computed in closed form, it is said that $\Omega$ is a \emph{proximable} function, and problem \eqref{eq:estimation} can be solved efficiently via proximal gradient descent (PGD):
\begin{equation}\label{eq:pgd}
    w^{(k+1)} =\mathrm{prox}^{ \lambda\alpha^{(k)}}_\Omega (w^{(k)} - \frac{\alpha^{(k)}}{|\mathcal{B}_j|}\sum_{i \in \mathcal{B}_j}(\nabla_w \ell_i(w^{(k)})))
\end{equation}
where $\alpha^{(k)}$ is a step size sequence satisfying $\alpha^{(k)} < 1/L$, where $L$ is the Lipschitz smoothness parameter of the empirical risk (aggregate loss component of the cost function). In fact, for $\alpha^{(k)} < 1/L$, it holds that 
$w_j^{(k)} \xrightarrow[k \to \infty]{} w^\ast(\lambda, \mathcal{B}_j).$
The PGD step \eqref{eq:pgd} is the composition of a gradient step with the prox operator, and the iteration is frequently split in two steps, yielding the equivalent \emph{forward-backward} iterations:
\begin{subequations}\label{eq:forward-backward}
\begin{align}
    w_f^{(k)}   = &
    F_{\mathcal{B}}^{\alpha^{(k)}} ( w^{(k)} ) 
    \triangleq w^{(k)} -   \frac{\alpha^{(k)}}{|\mathcal{B}_j|} \sum_{i \in \mathcal{B}} 
    \nabla_w \ell_i (w^{(k)} )
    \\
    w^{(k+1)} = &
    \mathrm{prox}^{ \lambda\alpha^{(k)}}_\Omega  ( w_f^{(k)} )
\end{align}
\end{subequations}

Moreover, for $\alpha \in (0, 1/L]$ the optimality condition holds:
\begin{equation}\label{eq:optimality_condition}
    w^\ast(\lambda, \mathcal{B}) = \mathrm{prox}^{ \lambda\alpha}_\Omega (F_{\mathcal{B}}^\alpha  (w^\ast(\lambda, \mathcal{B}))).
\end{equation}

\section{Computing the Hyper-gradient}\label{sec:hypergradient}

The condition in \eqref{eq:optimality_condition} establishes optimality w.r.t. the weight vector, but not w.r.t. the hyperparameter $\lambda$. To optimize over $\lambda$, we leverage the \emph{forward-mode} gradient computation described by \cite{franceschi2017forward} in this section. The condition for $\lambda^\ast$ being a stationary point for the optimization in \eqref{eq:bilevel} is:
\begin{equation}
    \sum_{j\in\mathcal{V}} \nabla_\lambda \ell_j^{\mathrm{VAL}}(w^\ast(\lambda^\ast, \mathcal{B}_j)) = 0.
\end{equation}
The hyper-gradient can be written using the chain rule as 
\begin{equation}
    \nabla_\lambda \ell_j^{^{\mathrm{VAL}}}\hspace{-1mm}(w^\ast(\lambda, \mathcal{B}))
    = 
    \Big(
        \frac{\partial w^\ast(\lambda, \mathcal{B})}
             {\partial \lambda}
    \Big)^{\hspace{-1mm}\top}
    \nabla_w \ell_j^{^{\mathrm{VAL}}}\hspace{-1mm}( w^\ast(\lambda, \mathcal{B}) )
    ,
\end{equation}
where the argument of $^\top$ is the derivative (Jacobian) matrix (column vector if $\lambda$ is scalar). In the sequel, we leverage the technique in \cite{franceschi2017forward} to compute the latter.


Consider a generic iterative algorithm, whose $t$-th iterate is $s_t \in \mathbb{R}^P$, and a hyperparameter vector $\lambda \in \mathbb{R}^D$. The $t$-th iteration can be expressed as:
$
s_t = \mathcal{M}_t(s_{t-1}, \lambda),
$
where 
$$
\mathcal{M}_t: (\mathbb{R}^P \times \mathbb{R}^D) \to \mathbb{R}^P 
$$
is a smooth mapping that represents the operation performed at the latter. The following equation \cite[eq. (13)]{franceschi2017forward} is fulfilled by the iterates $s_t$:
\begin{equation}\label{eq:forward_gradient_generic}
    \frac{d s_t}{d\lambda} = 
    \frac{\partial \mathcal{M}_t(s_{t-1}, \lambda)}{\partial s_{t-1}}
    \frac{d s_{t-1}}{d \lambda} + 
    \frac{\partial \mathcal{M}_t(s_{t-1}, \lambda)}{\partial \lambda}
\end{equation}

In the case of PGD, the mapping $\mathcal{M}_k$ is the composition $\mathrm{prox}_\Omega^{\lambda\alpha(k)} \circ F_\mathcal{B}^{\alpha(k)}$ [cf. \eqref{eq:forward-backward}]. For simplicity, we will consider in the sequel a constant step size $\alpha^{(k)} = \alpha$ for PGD, so that $\mathcal{M}_k = \mathcal{M} = \mathrm{prox}_\Omega^{\lambda\alpha} \circ F_\mathcal{B}^{\alpha}$, and
\begin{equation}\label{eq:forward_gradient_iterate}
    \frac{d w^{(k+1)}}{d\lambda} = 
    A(w_f^{(k)})
    \frac{\partial F_\mathcal{B}^\alpha(w^{(k)})}{\partial w^{(k)}}
    \frac{d w^{(k)}}{d\lambda} + 
    B(w_f^{(k)})
\end{equation}
\begin{equation}
    \text{where }  
    A(w_f)     \triangleq 
    \frac{\partial (\mathrm{prox}_\Omega^{\lambda\alpha})(w_{f})}{\partial w_{f}},
    B(w_f)    \triangleq 
    \frac{\partial (\mathrm{prox}_\Omega^{\lambda\alpha})(w_{f})}{\partial \lambda}.
\end{equation}

The derivations so far have followed a path common to \cite{franceschi2018bilevel}, where an approximation to the hypergradient is computed by reverse-mode gradient computation \cite{franceschi2017forward}. However, differently to this work, in our approach we identify a fixed point equation for the derivatives at the convergence point of PGD:
\begin{equation}\label{eq:forward_gradient_fixed_point}
    \frac{d w^\ast(\lambda, \mathcal{B})}{d\lambda} 
    =
    A(w_f^\ast)
    \frac{\partial F_\mathcal{B}^\alpha(w^\ast(\lambda, \mathcal{B}))}{\partial w^\ast(\lambda, \mathcal{B})}
    \frac{d w^\ast(\lambda, \mathcal{B})}{d\lambda} 
    + 
    B(w_f^\ast)
\end{equation}
where $w_f^\ast \triangleq F_\mathcal{B}^\alpha(w^\ast(\lambda, \mathcal{B}))$; if the linear equation has a solution, it can be expressed in closed form as $\frac{d w^\ast(\lambda, \mathcal{B})}{d\lambda} =
Z_\mathcal{B}(w^\ast(\lambda, \mathcal{B}))$, where
\begin{equation} \label{eq:derivative_Z}
Z_\mathcal{B}(w^\ast(\lambda, \mathcal{B}))
\triangleq
\left( \bm I -
A(w_{f}^\ast)
    \frac{\partial F_\mathcal{B}^\alpha(w^\ast(\lambda, \mathcal{B}))}{\partial w^\ast(\lambda, \mathcal{B})}
\right)^{-1}
B(w_{f}^\ast).
\end{equation}
\remove{
\begin{equation} \label{eq:derivative_Z}
\begin{split}
\frac{d w^\ast(\lambda, \mathcal{B})}{d\lambda} =
Z_\mathcal{B}(w^\ast(\lambda, \mathcal{B})) 
\\
\triangleq
\left( \bm I -
\frac{\partial (\mathrm{prox}_\Omega^{\lambda\alpha})(w_{f}^\ast)}{\partial w_f^\ast}
    \frac{\partial F_\mathcal{B}^\alpha(w^\ast(\lambda, \mathcal{B}))}{\partial w^\ast(\lambda, \mathcal{B})}
\right)^{-1}
\frac{\partial (\mathrm{prox}_\Omega^{\lambda\alpha})(w_{f}^\ast)}{\partial \lambda}.
\end{split}
\end{equation}
}

\subsection{Hyper-gradient descent (HGD)}

If the iterates \vspace{-0.5cm}
\begin{equation} \label{eq:alg_hgd}
\begin{split}
    \lambda^{(k+1)} := \Big[ 
        \lambda^{(k)} - \frac{\beta^{(k)}}{|\mathcal{V}|} \times 
        \\
        \sum_{j \in \mathcal{V}}
        \left(Z_\mathcal{B}(w^\ast(\lambda^{(k)}, \mathcal{B}_j)) \right)^\top
        \nabla_w \ell_j^{\mathrm{VAL}}(w^\ast(\lambda^{(k)}, \mathcal{B}_j))       
    \Big]_+
    \end{split}
\end{equation}
(where $[\cdot]_+$ denotes projection onto the positive orthant) are executed,
with an appropriate step size sequence $\beta^{(k)}$, the sequence $\lambda^{(k)}$ will converge to a stationary point of \eqref{eq:bilevel}.

\textbf{Remark.} Existence of $Z_{\mathcal{B}}(\cdot)$ requires the prox operator to be smooth. However, important estimation problems such as Lasso regression rely on non-smooth prox operators. In the next section, a slight modification of the hyper-gradient descent is proposed in order to deal with those problems.

\section{Non-smooth prox operators} \label{sec:nonsmooth}

In this section, we propose the hyper-subgradient descent method, and its extension for large datasets, namely, the online hyper-subgradient descent (OHSD) method.

If the prox operator is nonsmooth, its derivatives may not exist at all points, and thus $Z_j(w^\ast(\lambda, \mathcal{B}_j))$ may not be computable. One can instead compute a valid subderivative (which will be denoted by $\tilde Z_j(w^\ast(\lambda, \mathcal{B}_j))$) by replacing the derivatives of the prox operator with the corresponding subderivatives.

If $Z_j(w^\ast(\lambda, \mathcal{B}_j))$ is replaced in \eqref{eq:alg_hgd} with $\tilde Z_j(w^\ast(\lambda, \mathcal{B}_j))$, the resulting algorithm will be termed hereafter as \emph{hyper-subgradient descent (HSGD)}.

The HSGD will be advocated in the next section to optimize the hyperparameters for several estimation problems with nonsmooth prox operators, namely Lasso and Group Lasso.
Before proceeding, some of the functions that have been presented before as generic functions, will be particularized to facilitate the readability of the derivations and algorithms.

\cmt{LS, $\Phi$, $r$, $Phi_j$ and $r_j$ for LOO, $\ell^{VAL}$}\begin{myitemize}
\myitem Regularized least-squares (LS) linear estimators such as Lasso use the loss function $\ell_i(w) = (y_i - x_i^\top w)^2$. 
    Consequently, the forward operator and its Jacobian are
    $$
    F_\mathcal{B}^\alpha(w) = w - \alpha( \bm\Phi_j w - r_j ), \quad \text{and}
    \quad \frac{\partial F_\mathcal{B}^\alpha(w)}{\partial w} = (\bm I - \alpha \bm\Phi_j),
    $$ where 
    $\bm \Phi_j := \frac{1}{|\mathcal{B}_j|} \sum_{i \in \mathcal{B}_j} x_i x_i^\top $, and 
    $r_j :=  \frac{1}{|\mathcal{B}_j|} \sum_{i \in \mathcal{B}_j} y_i x_i$.
    \myitem If the LOO validation scheme is chosen, then $\bm\Phi_j$ can be computed efficiently as 
    \begin{equation} \label{eq:phi_j_r_j} 
    \textstyle
        \bm\Phi_j := \frac{1}{N-1}(N \bm\Phi - x_i x_i^\top ), \quad
        r_j := \frac{1}{N-1}(N r - x_i y_i);
    \end{equation}
    \begin{equation} \label{eq:phi_r} 
    \textstyle
    \text{with}\;
        \bm\Phi \triangleq \frac{1}{N} \sum_{i \in \mathcal{V}} x_i x_i^\top, \quad
        r \triangleq \frac{1}{N} \sum_{i \in \mathcal{V}} x_i y_i.
    \end{equation}            
    \myitem If the validation error metric is $\ell_j^\textrm{VAL} = (y_j - x_j^\top w)^2$, then
    $$
    \nabla_w \ell_j^{\mathrm{VAL}}(w) = x_j ( x_j^\top w - y_j).
    $$
\end{myitemize}
The equations for particular cases of $\Omega(\cdot)$ will be presented after the HSGD 
algorithm.

\subsection{Hyper-subgradient descent (HSGD)}
Let $\tilde A_j(w_f)$ and $\tilde B_j(w_f)$ be valid subderivative (sub-Jacobian) matrices of $\mathrm{prox}_\Omega^{\lambda\alpha}(w_f)$ w.r.t. $w_f$ and $\lambda$, respectively. Then, a valid subderivative matrix of $w^\ast(\lambda, \mathcal{B}_j)$ with respect to $\lambda$ is [cf. \eqref{eq:derivative_Z}]
\begin{equation}\label{eq:subderivative_Z}
    \tilde Z_j(w^\ast(\lambda, \mathcal{B}_j)) 
    := 
    \left( \bm I -
    \tilde A_j (w_f^\ast)
    (\bm I - \alpha \bm\Phi_j)
    \right)^{-1}
    \tilde B_j(w_f^\ast);
\end{equation}
where $w_f^\ast := F_\mathcal{B}^\alpha (w^\ast(\lambda, \mathcal{B}))$; and the HSGD iterates can be written as \vspace{-0.5cm}
\begin{equation}\label{eq:alg_hsgd}
\begin{split}
    \lambda^{(k+1)} := \Big[ 
        \lambda^{(k)} - \beta^{(k)} \times 
        \\ 
        \sum_{j \in \mathcal{V}}
        \left(\tilde Z_j(w^\ast(\lambda^{(k)}, \mathcal{B}_j)) \right)^\top
        x_j(x_j^\top w^\ast(\lambda^{(k)}, \mathcal{B}_j)- y_j)       
    \Big]_+
\end{split}
\end{equation}

\textbf{Remark.} The inverse at \eqref{eq:subderivative_Z} will not exist if $\bm \Phi_j$ is rank-defficient. This happens when the model dimensionality $P$ is less than $N+1$, and may also happen when the input data $x_j$ have a high degree of colinearity. In such cases, the LS solution of the linear system can be used. Another option is to numerically approximate $\tilde Z(\cdot)$ by using an iterative algorithm based on the forward-gradient iteration at \eqref{eq:forward_gradient_iterate}.

\subsection{Application of HSGD to Lasso and Group Lasso}

Depending on the choice of the function $\Omega$, we obtain different regularized estimators, and associated prox operators and HSGD iterates. 

\subsubsection{Lasso}
    The regularizer is
        \begin{myitemize} 
            \myitem\cmt{reg fn}$\Omega(w) = \| w\|_1$; 
                \myitem\cmt{prox = soft thresholding} its prox operator is known as soft-thresholding
                $   S_{\alpha\lambda}(w) 
                    \triangleq 
                    \mathrm{prox}_{\|\cdot \|_1}^{\alpha\lambda}(w)  
                $  \cite{daubechies2004ISTA}, and the latter can be computed entrywise as
                    \begin{equation}\label{eq:prox_lasso}
                        [S_{\alpha\lambda}(w_f)]_n := [w_f]_n \left[ 1 - \frac{\alpha\lambda}{|[w_f]_n|}\right]_+.
                    \end{equation}
            \myitem\cmt{sub-Jacobian matrices} 
                The corresponding subderivatives $\tilde A (w_f) \in \mathbb{R}^{P\times P}$, and $\tilde B(w_f) \in \mathbb{R}^{P\times 1}$ are defined so that $\tilde A (w_f)$ is diagonal and
                \begin{subequations}   \label{eq:subderivatives_lasso}
                    \begin{align}
                        [\tilde A (w_f)]_{nn} = & 
                        \mathbb{1}\{|[w_f]_n| \geq {\alpha \lambda}\}
                        \\
                        [\tilde B(w_f)]_n = & 
                        \alpha\left(\mathbb{1}\{[w_f]_n \leq {-\alpha \lambda}\} - \mathbb{1}\{[w_f]_n \geq {\alpha \lambda}\} \right).
                    \end{align}
                \end{subequations}    
            \remove{ 
            \myitem The corresponding subderivative matrices are: 
                \begin{equation}
                    \tilde A (w_f):= \mathrm{Diag}(
                    [\mathbb{1}\{|[w_f]_1| \geq {\alpha \lambda}\}, 
                     \mathbb{1}\{|[w_f]_2| \geq {\alpha \lambda}\}, 
                                 \ldots, 
                     \mathbb{1}\{|[w_f]_P| \geq {\alpha \lambda}\} ]^\top),
                \end{equation}
                and $\tilde B(w_f) \in \mathbb{R}^P$ is defined so that 
                \begin{equation}
                    [\tilde B(w_f)]_n = \alpha\left(\mathbb{1}\{[w_f]_n \leq {-\alpha \lambda}\} - \mathbb{1}\{[w_f]_n \geq {\alpha \lambda}\} \right).
                \end{equation}
                }
        \end{myitemize}
        \vspace{-3mm}
    \subsubsection{Group Lasso} The regularizer depends on an a priori defined group structure.   
        \begin{myitemize} 
            \myitem\cmt{group structure} 
                With $P$ denoting the dimensionality of $w$, and $N_g$ the number of groups, let $\{ \mathcal{K}_1, \mathcal{K}_2, ... \mathcal{K}_{Ng}\}$ be a partition of $\{1, 2, ...,  P\}$. Let $[w]_{\mathcal{K}}$ denote the sub-vector of $w$ containing the components indexed by $\mathcal{K}$.
            \myitem\cmt{reg fn} The regularizer is $\Omega(w) = \|w\|_{2,1} \triangleq \sum_{g=1}^{N_g} \|w_{\mathcal{K}_g}\|_2$;
            \myitem\cmt{prox = group soft thresholding} its prox operator is known as multidimensional soft-thresholding $
                S^G_{\alpha\lambda}(w) 
                \triangleq 
                \mathrm{prox}_{\|\cdot\|_{2,1}}^{\alpha\lambda}(w)$ \cite{puig2011multidimensional}, and the latter can be computed group-wise as
                \begin{equation}\label{eq:prox_group_lasso}
                    [S^{G}_{\alpha\lambda}(w_f)]_\mathcal{K} = [w_f]_\mathcal{K} \left[ 1 - \frac{\alpha\lambda}{\|[w_f]_\mathcal{K}\|_2}\right]_+.
                \end{equation}
                \myitem\cmt{sub-Jacobians} With $\mathcal{K}(n)$ denoting the subset of the partition where $n$ belongs, the corresponding subderivative matrices $\tilde A (w_f) \in \mathbb{R}^{P\times P}$, and $\tilde B(w_f) \in \mathbb{R}^{P\times 1}$ are defined so that $\tilde A (w_f)$ is diagonal, and
                \begin{subequations}   \label{eq:subderivatives_group_lasso}
                     \begin{align}
                        [\tilde A (w_f)]_{nn} = & 
                        \mathbb{1}\{\|[w_f]_{\mathcal{K}(n)}\|_2 \geq {\alpha \lambda}\}
                        \\
                        [\tilde B(w_f)]_n = & 
                        \begin{cases}
                            -\alpha \frac{[w_f]_{\mathcal{K}(n)}}{\|[w_f]_{\mathcal{K}(n)}\|_2},
                            & \|[w_f]_{\mathcal{K}(n)}\|_2 \geq {\alpha \lambda}, 
                            \\
                            \;\;\; 0, & \|[w_f]_{\mathcal{K}(n)}\|_2 < {\alpha \lambda}.
                        \end{cases}
                    \end{align}
                \end{subequations}
                \remove{
                \begin{itemize}
                    \item $\tilde A (w_f):= \mathrm{BlkDiag}(
                        \mathbb{1}\{\|[w_f]_{\mathcal{K}_1}\|_2 \geq {\alpha \lambda}\} \bm I_{|\mathcal{K}_1|}, 
                        \mathbb{1}\{\|[w_f]_{\mathcal{K}_2}\|_2 \geq {\alpha \lambda}\}\bm I_{|\mathcal{K}_2|}, 
                                     \ldots, 
                        \mathbb{1}\{\|[w_f]_{\mathcal{K}_{N_g}}\|_2 \geq {\alpha \lambda} \bm I_{|\mathcal{K}_{N_g}|}\})$
                    \item $\tilde B(w_f) \in \mathbb{R}^P$ is defined so that 
                    $$
                        [\tilde B(w_f)]_\mathcal{K} = \alpha\left(\mathbb{1}\{_\|[w_f]\mathcal{K}\| \leq {-\alpha \lambda}\} - \mathbb{1}\{\|[w_f]_\mathcal{K}\| \geq {\alpha \lambda}\} \right) \bm 1_{|\mathcal{K}|}.
                    $$
                \end{itemize}
                }
        \end{myitemize}
        \acom{Weighted Lasso is hidden}
        \remove{
    \myitem Weighted Lasso: 
        \begin{myitemize} 
            \item The regularization term in this case is $\bm \lambda^\top \bm \Omega(w) = \sum_n \lambda_n |[w]_n|$. 
            \item The hyperparameter $\bm \lambda$ is a vector of length $P$. 
            \item The regularizer $\bm \Omega(w)$ is a vector function: $\mathbb{R}^P \to \mathbb{R}^P$
            \item The weighted soft-thresholding operator can be computed entrywise as:
                \begin{equation}
                    [S^W_{\alpha \bm \lambda}(w_f)]_n := [w_f]_n \left[ 
                    1 - \frac{\alpha\lambda_n}{|[w_f]_n|} \right]_+.
                \end{equation}
            \item The subderivative matrices are: 
                \begin{itemize}
                     \item $\tilde A (w_f):= \mathrm{Diag}(
                        [\mathbb{1}\{|[w_f]_1| \geq {\alpha \lambda_1}\}, 
                         \mathbb{1}\{|[w_f]_2| \geq {\alpha \lambda_2}\}, 
                                     \ldots, 
                         \mathbb{1}\{|[w_f]_P| \geq {\alpha \lambda_P}\} ]^\top)$
                    \item $\tilde B(w_f) \in \mathbb{R}^{P \times P}$ is a diagonal matrix defined so that
                    $$
                        [\tilde B(w_f)]_{nn} = \alpha\left(\mathbb{1}\{[w_f]_n \leq {-\alpha \lambda_n}\} - \mathbb{1}\{[w_f]_n \geq {\alpha \lambda_n}\} \right).
                    $$
                \end{itemize}
        \end{myitemize}
        }

\newcommand{\Lasso}[1]{\textcolor{blue}{#1}}
\newcommand{\GLasso}[1]{\textcolor{lightgreen}{#1}}
\begin{algorithm}
    \caption{
        Hyper-subgradient descent for \Lasso{Lasso} or \GLasso{Group Lasso}
        }
    \label{alg:hsgd-lasso}
    \textbf{Input:} 
        $\{x_i, y_i\}_{i=1}^N ,\; \{\beta^{(k)}\}_k ,\; \lambda^{(1)}$ \\
    \textbf{Output:} 
        $\lambda^\ast$
    \begin{algorithmic}[1] 
        \State{Compute $\bm \Phi$, $r$ via \eqref{eq:phi_r}}
        \State{$\alpha = 1/\rho(\bm \Phi)$} 
        \For {$k=1, 2, \ldots $} (until convergence) 
            \For {$j=1, \ldots, N $}
                \State{Compute $\bm \Phi_j$, $r_j$ via \eqref{eq:phi_j_r_j}}
                \For {$m=1, 2, \ldots $ } (until convergence) \Comment{PGD}
                    \State{$w_f^{(m)} = w^{(m-1)} - \alpha( \bm\Phi_j w^{(m-1)} - r_j )$}
                    \State{Compute $w^{(m)}$ via \Lasso{\eqref{eq:prox_lasso}}} or \GLasso{\eqref{eq:prox_group_lasso}}
                    \EndFor
                \State{Compute $\tilde A_j(w_f^\ast), \tilde B_j(w_f^\ast)$ via \Lasso{\eqref{eq:subderivatives_lasso}}  or \GLasso{\eqref{eq:subderivatives_group_lasso}}}
                \State{Compute $\tilde Z_j(w^\ast(\lambda^{(k)}, \mathcal{B}_j))$ via \eqref{eq:subderivative_Z}}
            \EndFor
            \State{Update $\lambda^{(k+1)}$ via \eqref{eq:alg_hsgd}}
        \EndFor
    \end{algorithmic}
\end{algorithm}%
The HSGD algorithm applied to Lasso and Group Lasso is summarized in the \textbf{Algorithm} \ref{alg:hsgd-lasso}.
The approach in this paper can be extended also to other estimators with proximable regularizers, particularly several generalizations of Lasso such as Weighted Lasso and Fused Lasso, which are left out of the scope of this article for space constraints.

\section{Approximate algorithms} \label{sec:approximate}

This section presents two approximations that improve the efficiency of HSGD.

\subsection{Online Hyper-subgradient Descent (OHSGD)}
To avoid having to evaluate $w^\ast(\lambda, \mathcal{B}_j)$ for all $j$ in each iteration of HSGD, the online optimization technique is applied here, which consists in doing a gradient descent iteration per $j$, using the corresponding contribution to the subgradient (also known as stochastic subgradient): 
\begin{subequations}\label{eq:ohsgd}
\begin{align}
    j(k) := &
    k \mod \vert \mathcal{V} \vert
    \\
    w^{(k)} := &
    w^\ast (\lambda^{(k)}, \mathcal B_{j(k)})
    \\
    \lambda^{(k+1)} := & 
        \Big[ \lambda^{(k)} - \beta^{(k)} \times \nonumber
        \\ &
        \Big(\tilde Z_{j(k)}(w^{(k)}) \Big)^\top
        x_{j(k)}(x_{j(k)}^\top w^{(k)}- y_{j(k)})       
        \Big]_+
\end{align}
\end{subequations}

To save computation, the instance of PGD that calculates $ w^\ast (\lambda^{(k)}, \mathcal B_{j(k)})$ should be initialized at $w^{(k-\vert \mathcal{V} \vert)}$ if $k>\vert \mathcal{V} \vert$.

\subsection{OHSGD with inexact weight vector}
The algorithm proposed in the previous section requires to evaluate $w^\ast (\lambda^{(k)}, \mathcal B_{j(k)})$. The iterates produced by PGD converge to the exact optimizer, but in practice one has to stop the inner loop after a certain stopping criterion is met. Clearly, there is a trade-off between the number of iterations $m(k)$ in the $k$-th (inner) loop and the suboptimality of its final iterate, $\| w_j^{(m(k))} - w^\ast(\lambda, \mathcal{B}_{j(k)})\|$. 

Even if one is interested in a very precise approximation of $(\lambda^\ast ,\{w^\ast(\lambda^\ast, \mathcal{B}_j)\}_{j\in \mathcal{V}})$, most of the times PGD is run to evaluate $w^\ast(\lambda^{(k)}, \mathcal{B}_{j(k)})$ for $\lambda^{(k)}$ far away from $\lambda^\ast$, and $w^\ast(\lambda^{(k)}, \mathcal{B}_{j(k)})$ is only used to compute the hypergradient. It is well known that when applying gradient methods, using coarsely approximated (hyper) gradients before getting close to a stationary point usually does not hinder the convergence, and may significantly alleviate computation. Even if the number of hyper-gradient steps required for converge increases, the computation savings in the inner loop usually yield a faster overall convergence. In addition, if the the prox operator is computationally heavy, fast (inexact) approximations of the prox operator also lower the complexity per iteration (inexact PGD method) \cite{schmidt2011convergence}.


\acom{Memoryless approx. and Dynamic sections are hidden}
\remove{
\subsection{Memoryless approximation to the OHSGD with inexact weights} \label{ss:memoryless}

Even though we use the inexact weight vector, the warm start requires storing $N$  weight vectors (as many as data points), which may exceed the available memory if $N$ is very large (big data). 

An approximation that possibly reduces computation is to set the initialization point for the inner loop in the $k$-th hypergradient step (the one approximating $ w^\ast (\lambda^{(k)}, \mathcal B_{j(k)})$) as $w^{(k-1)}$.

\acom{TODO: evaluate this empirically.} 

\acom{This method may even not converge to a neighbourhood of the optimal hyperparameter vector, but it is important to discuss this approach as it can be taken as a preliminary step to the dynamic setting, where we track a hyperparameter that changes slowly/smoothly with time}

}
\section{Numeric tests}\label{sec:numeric}

For the two experiments in this section, data are generated so that the inputs $x_i \in \mathbb{R}^{100}$ are i.i.d., and $y_i := w_{\mathrm{true}}^\top x_i + \epsilon_i$, with $ w_{\mathrm{true}}$ being a 10-sparse vector, and $\epsilon_i$ generated i.i.d. so that $y_i$ has a signal-to-noise ratio (SNR) of $0.3$. The train-and-validate set contains 200 samples. A test set is generated with the same model and 2000 samples.

\begin{figure}
    \centering
    \includegraphics[width=0.9\columnwidth]{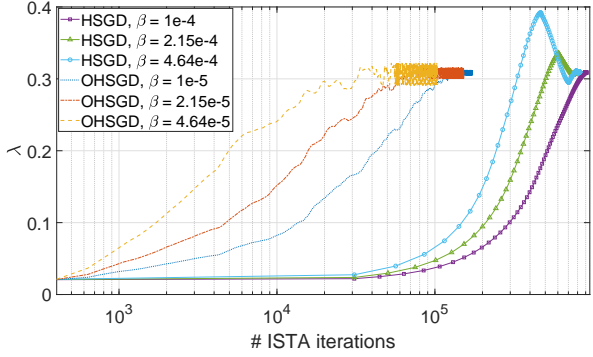}
    \caption{\small{Iterates of HSGD and OHSGD for different values of $\beta$}}
    \label{fig:online_vs_offline}
\end{figure}

The first experiment is run in order to visually compare in Fig. \ref{fig:online_vs_offline} the convergence rates of HSGD and OHSGD with different constant stepsizes $\beta^{(k)} = \beta$, in terms of the number of PGD/ISTA iterations executed before producing a given value of $\lambda$. The tolerance to stop the inner loop is set to 1e-3.

\begin{figure}
    \centering
    \includegraphics[width=\columnwidth]{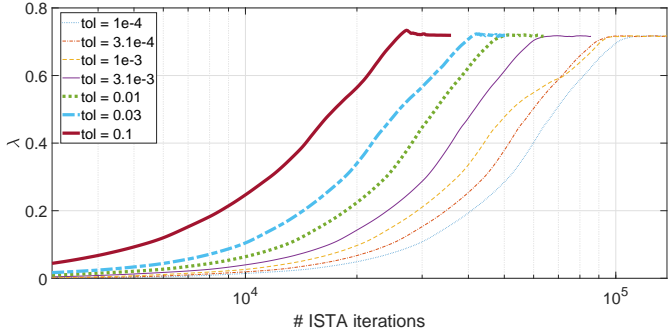}
    \caption{\small OHSGD iterates for $\beta = 6e-5$, and different values of the tolerance to stop PGD/ISTA. }
    \label{fig:conv_tol}
\end{figure}

\begin{figure}
    \centering
    \includegraphics[width=0.7\columnwidth]{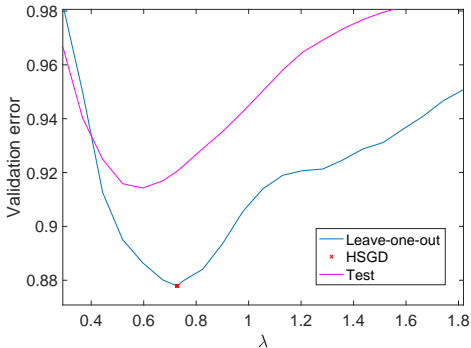}
    \caption{\small Validation error of LOO, Test, and solution generated by OHSGD (experiment 2)}
    \label{fig:validation_error}
\end{figure}

The second experiment consists in evaluating the convergence rate of OHSGD with inexact weight vectors within a scale of coarser-finer approximate values of the optimal solution of \eqref{eq:optimality_condition}. Fig. \ref{fig:conv_tol} shows the value of the $\lambda$ iterates (averaged over the last $N_{train}$ to show a stable value, since online iterates hover around the optimizer) against the number of PGD (ISTA) iterations. The PGD loop is stopped when the distance between 0 and subgradient of the training loss is smaller than tol. To confirm the optimality of $\lambda^\ast$, Fig. \ref{fig:validation_error} shows the LOO and test error curves for a grid of values for $\lambda$.

The results show that approximate weights as with a subgradient tolerance as coarse as 0.1 still allow convergence of $\lambda$ to $\lambda^\ast$, and the computation is significantly reduced with respect to instances of OHSGD that calculate the weights more exactly.

\textbf{Concluding remarks:} In this paper, the (hyper)gradient of the validation error w.r.t. the hyperparameters has been derived for estimators with non-smooth regularizers exploiting the structure of PGD. An algorithm has been developed (with an online variant) to optimize hyperparameters for Lasso and Group Lasso. Actually, this approach is flexible enough to accomodate any convex, proximable regularization term. 

\if\editmode1 
\onecolumn
\printbibliography
\else
\bibliography{\bibfilenames}

\begin{thebibliography}{10}

\bibitem{bergstra2011hyper}
James~S Bergstra, R{\'e}mi Bardenet, Yoshua Bengio, and Bal{\'a}zs K{\'e}gl,
\newblock ``Algorithms for hyper-parameter optimization,''
\newblock in {\em Proc. Advances Neural Inf. Process. Syst.}, 2011, pp.
  2546--2554.

\bibitem{homrighausen2014leave}
Darren Homrighausen and Daniel~J McDonald,
\newblock ``Leave-one-out cross-validation is risk consistent for lasso,''
\newblock {\em Machine learning}, vol. 97, no. 1-2, pp. 65--78, 2014.

\bibitem{bergstra2012random}
James Bergstra and Yoshua Bengio,
\newblock ``Random search for hyper-parameter optimization,''
\newblock {\em J. Mach. Learn. Res.}, vol. 13, no. Feb, pp. 281--305, 2012.

\bibitem{li2018hyperband}
Lisha Li, Kevin Jamieson, Giulia DeSalvo, Afshin Rostamizadeh, and Ameet
  Talwalkar,
\newblock ``Hyperband: A novel bandit-based approach to hyperparameter
  optimization,''
\newblock {\em J. Mach. Learn. Res.}, vol. 18, no. 1, pp. 6765--6816, 2018.

\bibitem{klein2017fast}
Aaron Klein, Stefan Falkner, Simon Bartels, Philipp Hennig, and Frank Hutter,
\newblock ``Fast bayesian optimization of machine learning hyperparameters on
  large datasets,''
\newblock in {\em Artificial Intelligence and Stat.}, 2017, pp. 528--536.

\bibitem{pedregosa2016approximate}
Fabian Pedregosa,
\newblock ``Hyperparameter optimization with approximate gradient,''
\newblock {\em arXiv preprint arXiv:1602.02355}, 2016.

\bibitem{monti2018adaptive}
Ricardo~P Monti, Christoforos Anagnostopoulos, and Giovanni Montana,
\newblock ``Adaptive regularization for lasso models in the context of
  nonstationary data streams,''
\newblock {\em Stat. Analysis and Data Mining: The ASA Data Science Journal},
  vol. 11, no. 5, pp. 237--247, 2018.

\bibitem{franceschi2017forward}
Luca Franceschi, Michele Donini, Paolo Frasconi, and Massimiliano Pontil,
\newblock ``Forward and reverse gradient-based hyperparameter optimization,''
\newblock in {\em Proc. Int. Conf. Mach. Learn.}, 2017, vol.~70, pp.
  1165--1173.

\bibitem{franceschi2018bilevel}
Luca Franceschi, Paolo Frasconi, Saverio Salzo, Riccardo Grazzi, and
  Massimiliano Pontil,
\newblock ``Bilevel programming for hyperparameter optimization and
  meta-learning,''
\newblock in {\em Proc. Int. Conf. Mach. Learn.}, 2018, pp. 1568--1577.

\bibitem{lorraine2018stochastic}
Jonathan Lorraine and David Duvenaud,
\newblock ``Stochastic hyperparameter optimization through hypernetworks,''
\newblock {\em arXiv preprint arXiv:1802.09419}, 2018.

\bibitem{wang2018approximateloo}
Shuaiwen Wang, Wenda Zhou, Arian Maleki, Haihao Lu, and Vahab Mirrokni,
\newblock ``Approximate leave-one-out for high-dimensional non-differentiable
  learning problems,''
\newblock {\em arXiv preprint arXiv:1810.02716}, 2018.

\bibitem{maclaurin2015gradient}
Dougal Maclaurin, David Duvenaud, and Ryan Adams,
\newblock ``Gradient-based hyperparameter optimization through reversible
  learning,''
\newblock in {\em International Conference on Machine Learning}, 2015, pp.
  2113--2122.

\bibitem{parikh2014proximal}
N.~Parikh and S.~Boyd,
\newblock ``Proximal algorithms,''
\newblock {\em Found. Trends Optim.}, vol. 1, no. 3, pp. 127--239, 2014.

\bibitem{daubechies2004ISTA}
Ingrid Daubechies, Michel Defrise, and Christine De~Mol,
\newblock ``An iterative thresholding algorithm for linear inverse problems
  with a sparsity constraint,''
\newblock {\em Communications on Pure and Applied Mathem.}, vol. 57, no. 11,
  pp. 1413--1457, 2004.

\bibitem{puig2011multidimensional}
Arnau~Tibau Puig, Ami Wiesel, Gilles Fleury, and Alfred~O Hero,
\newblock ``Multidimensional shrinkage-thresholding operator and group lasso
  penalties,''
\newblock {\em IEEE Signal Processing Letters}, vol. 18, no. 6, pp. 363--366,
  2011.

\bibitem{schmidt2011convergence}
Mark Schmidt, Nicolas~L Roux, and Francis~R Bach,
\newblock ``Convergence rates of inexact proximal-gradient methods for convex
  optimization,''
\newblock in {\em Proc. Advances Neural Inf. Process. Syst.}, 2011, pp.
  1458--1466.

\end{thebibliography}
\fi
\end{document}